\documentclass[letterpaper, 10 pt, conference]{ieeeconf}  

\IEEEoverridecommandlockouts                              

\usepackage{graphics} 
\usepackage{epsfig} 
\usepackage{mathptmx} 
\usepackage{times} 
\usepackage{amsmath} 
\usepackage{amssymb}  
\usepackage{algorithm}
\usepackage{algorithmic}
\usepackage{gensymb}
\usepackage{multirow}
\usepackage{subcaption}
\usepackage{cite}
\usepackage{tikz}
\usetikzlibrary{shapes,arrows,chains}
\usepackage{pgfplots}

\makeatletter
\let\NAT@parse\undefined
\makeatother
\usepackage{hyperref}

\usepackage{color}

\title{\LARGE \bf
A Vision-Centric Approach for Static Map Element Annotation
}

\author{\normalsize Jiaxin Zhang$^{1}$, Shiyuan Chen$^{1}$, Haoran Yin$^{1}$, Ruohong Mei$^{1}$, Xuan Liu$^{2}$, Cong Yang$^{1\dag}$, Qian Zhang$^{3}$ and Wei Sui$^{1}$}%

\begin{document}

\twocolumn[{
\renewcommand\twocolumn[1][]{#1}
\maketitle

\begin{center}   
    \centering
    \includegraphics[width=1.0\linewidth]{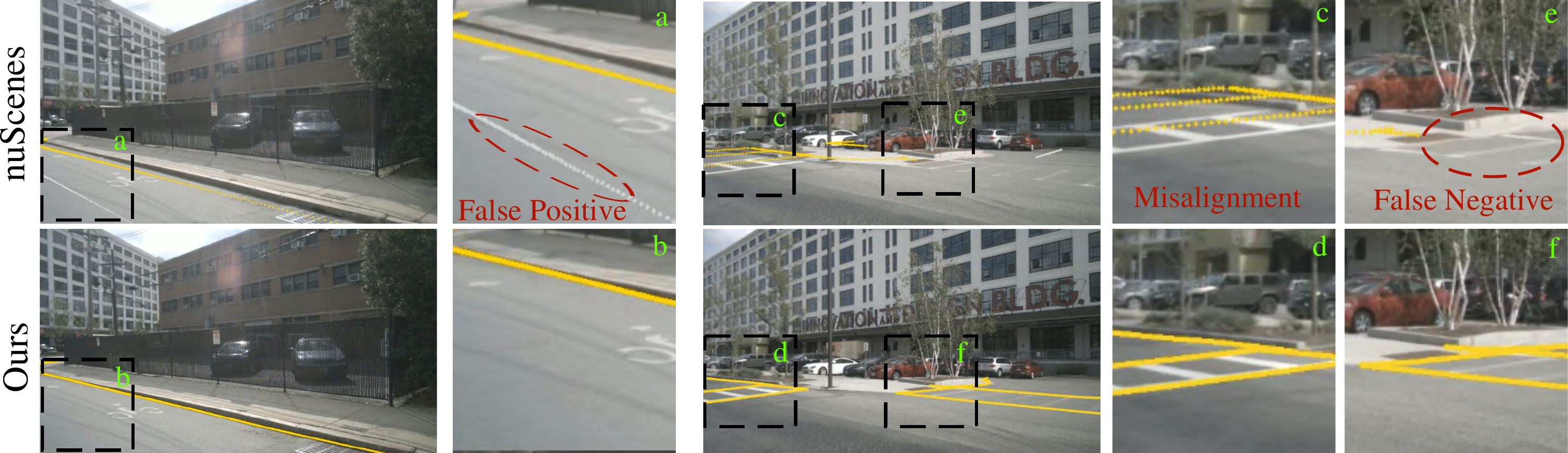}
    
    \captionof{figure}{\small Reprojection consistency and accuracy comparison. The top and bottom lines show HD map reprojection images and zoom-in details of nuScenes and our proposed method, respectively. The yellow dots represent road teeth, and the white dots represent lane dividers. The nuScenes HD Map has some inconsistent elements with respect to the actual road environments, including false positive (FP) and false negative (FN) road element annotation. For example, the image shows no lane marking between the bicycle and vehicle lane, but the HD Map indicates a lane divider (a,b). The image shows ped-crossing marking but no HD Map marking in the corresponding area (e, f). In contrast, the HD Map from our proposed method shows better reprojection accuracy (c, d) and consistency.
    }
   \label{fig:reprojection}
\end{center}%
}]

\let\thefootnote\relax\footnotetext{
$\dag$ Corresponding author: Cong Yang (cong.yang@suda.edu.cn).
Affiliation:
$^{1}$ Ecology and Innovation Center of Intelligent Driving, Soochow University, Suzhou, China.
$^{2}$ School of Information Science and Technology, Northeast Normal University, Changchun, China.
$^{3}$ Horizon Robotics, Beijing, China. The code and dataset are available at \href{https://github.com/manymuch/CAMA}{https://github.com/manymuch/CAMA}. 
This work was supported in part by the Natural Science Foundation of the Jiangsu Higher Education Institutions of China (22KJB520008); in part by the Research Fund of Horizon Robotics (H230666); and in part by the Jiangsu Policy Guidance Program, International Science and Technology Cooperation, The Belt and Road Initiative Innovative Cooperation Projects (BZ2021016).
}

\begin{abstract}
The recent development of online static map element (a.k.a. HD Map) construction algorithms has raised a vast demand for data with ground truth annotations. However, available public datasets currently cannot provide high-quality training data regarding consistency and accuracy. To this end, we present \underline{CAMA}: a vision-centric approach for \underline{C}onsistent and \underline{A}ccurate \underline{M}ap \underline{A}nnotation. Without LiDAR inputs, our proposed framework can still generate high-quality 3D annotations of static map elements. Specifically, the annotation can achieve high reprojection accuracy across all surrounding cameras and is spatial-temporal consistent across the whole sequence. We apply our proposed framework to the popular nuScenes dataset to provide efficient and highly accurate annotations. Compared with the original nuScenes static map element, models trained with annotations from CAMA achieve lower reprojection errors (e.g., 4.73 vs. 8.03 pixels).
\end{abstract}

\section{INTRODUCTION}
\label{sec:intro}

\begin{figure*}[t!]
    \includegraphics[width=1\textwidth]{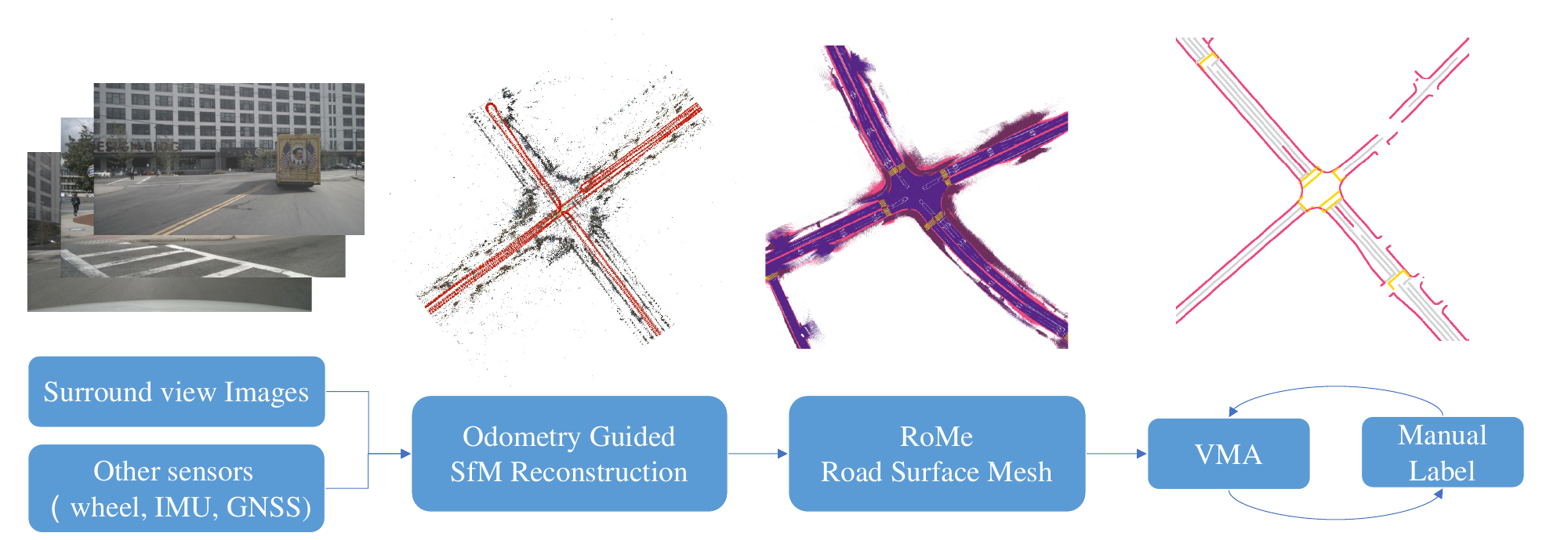}
  \caption{Illustration of our proposed reconstruction and annotation pipeline. The surround images and auxiliary sensor data are fed into our proposed odometry-guided SfM to obtain highly accurate ego vehicle poses and sparse 3D points. A road surface mesh reconstruction called RoMe is applied to build dense 3D road surfaces with semantic labels. Finally, a vectorized map annotation (VMA) System is applied to produce a 3D HD map required by the perception algorithm as training data.}
  \label{fig:whole_pipeline}
  \vspace{-1em}
\end{figure*}

The technical stack of perception algorithms for self-driving has recently been transforming from rule-based to data-driven methods~\cite{rao2018deep, gupta2021deep, zhou2023corner}.
Notably, the online high-definition map (HD Map) construction is becoming the mainstream of LiDAR-based and vision-centric bird-eye view (BEV) perception~\cite{ma2022vision}. These methods train neural networks that take single or surrounding camera images as input and then estimate the surrounding environments directly in BEV and 3D space. The spatial transform from the perspective view to BEV is usually conducted in the neural network models explicitly~\cite{garnett20193d, philion2020lift, reiher2020sim2real} or implicitly~\cite{li2022bevformer, liu2022petr, chen2022efficient}. These data-driven methods for perception algorithms drastically boost the advance of self-driving applications. It has several advantages, including less engineering efforts in corner cases debugging thanks to a data-driven closed-loop mechanism; better generalizability for various driving environments from the highway to city roads; better handling of occlusion and extreme illuminance conditions through temporal~\cite{huang2022bevdet4d, li2023bevstereo, huang2022bevpoolv2, wang2022mv} and raw image inputs~\cite{diamond2021dirty} in an end-to-end training manner.

Existing online HD map construction algorithms usually require high-quality and diverse labeled training data~\cite{zhou2022cross}. Accordingly, plenty of public datasets now provide annotation directly in 3D space. The available annotations can be roughly divided into HD-map-based and depth-reprojection-based~\cite{yan2022once}. For instance, the nuScenes dataset~\cite{caesar2020nuscenes} provides HD Map alongside ego-poses. The HD-MapNet~\cite{li2022hdmapnet} is one of the pioneers that utilizes the nuScenes HD-Map as ground truth and trains a neural network to predict road elements directly in BEV space. Persformer~\cite{chen2022persformer} uses the depth-reprojection method to generate 3D lane annotation. Specifically, the LiDAR point clouds are projected into image space. Combined with 2D lane segmentation, 3D lane point clouds for each frame are generated through reprojection.

However, there are several challenges with these annotation methods regarding accuracy and consistency. We argue that annotating 3D road elements should royally reflect the actual world environments. In detail, we propose two aspects for analyzing: consistency and geometric accuracy. Fig.~\ref{fig:reprojection} (top line) shows an example that the default HD Map from the nuScenes dataset can not provide accurate information in these two aspects. The consistency emphasizes the correspondence between the 3D annotations and the 2D images. For example, in some areas, the lane divider between the vehicle and bicycle lane is provided in the HD Map. At the same time, the actual image shows no lane marking in the corresponding area. The geometric accuracy is reflected in the reprojection of the HD Map into original images. The reprojected road teeth (yellow dots) deviate from the actual road teeth in the image. The main reason is that the HD Map provided by the dataset does not have elevation information, and the ego-motion is not accurate with respect to the maps.

In light of these challenges, we propose \textbf{CAMA}: a vision-centric approach for \textbf{C}onsistent and \textbf{A}ccurate \textbf{M}ap \textbf{A}nnotation (see Fig.~\ref{fig:whole_pipeline}). Our proposed CAMA is distinguished in three aspects: (1) We propose using a whole 3D reconstruction pipeline to get accurate camera motion and a sparse point cloud mainly from surround images. Thus, it can be applied to even low-cost self-driving platforms without equipping LiDAR. (2) A road surface mesh reconstruction algorithm~\cite{mei2023rome} is applied to reconstruct high-accuracy road surfaces. It can produce dense 3D road surfaces with both semantic and photometric information. (3) An auto map annotation tool~\cite{chen2023vma} is applied to extract the vectorized lane representation from the road surface. Consequently, as shown in Fig.~\ref{fig:reprojection} (bottom), CAMA achieves high consistency and geometric accuracy compared with nuScenes default HD Map. Succinctly, our main contributions are as follows:
\begin{itemize}
    \item We propose an efficient static element annotation framework, CAMA, for 3D road element annotation. The proposed CAMA can generate highly consistent and geometric accurate HD Map annotations. Through comprehensive experiments, we verified that such annotations can dramatically improve the accuracy and generalization of perception models in intelligent driving.
    \item To verify our proposed framework, we apply CAMA to the nuScene dataset and set up new HD Map annotations, namely nuScenes-CAMA. MapTR v2~\cite{liao2023maptrv2} is used as a benchmark model. Extensive experiments show that models trained on nuScenes-CAMA can produce more consistent and accurate estimations of the static map elements compared with default HD Map.
\end{itemize}

\section{RELATED WORKS}
\label{sec:related}

Existing driving datasets make the most effort in data collection, calibration, and manual annotations. However, recent BEV perception algorithms raise the demand for high-accuracy 3D road surface element annotation, which is hard to obtain by manual annotation only. To fulfill the requests for 3D road surface ground truth and BEV perception algorithm training, a pipeline from data collection, calibration, scene reconstruction, and annotation must be built and considered together. To the best of our knowledge, existing publicly available datasets~\cite{caesar2020nuscenes, chen2022persformer, li2023topology, yan2022once, chang2019argoverse, wilson2023argoverse} only meet a portion of such requirements.

HDMapNet~\cite{li2022hdmapnet} is one of the pioneers in predicting road elements using a neural network in BEV spaces. The BEV segmentation results from the networks are further processed into vectorized representation by post-processing. The authors proposed to use the HD Map provided by the nuScenes dataset for training. Following HDMapNet, MapTR \cite{liao2022maptr, liao2023maptrv2} further boosts the performance by improving the decoder and loss function modeling. To make a step towards end-to-end road structure understanding, LaneGAP \cite{liao2023lane}, and TopoNet \cite{li2023topology} regress the road topology structure directly. For a comprehensive development of the vision-centric BEV perception algorithm, we refer the readers to this survey \cite{ma2022vision}. All these methods use the same annotations provided by the nuScenes dataset. However, the provided HD Map lacks elevation information. As a result, the reprojection accuracy between the HD map and images is not guaranteed. Fig.~\ref{fig:reprojection} (top) shows the misalignment of the image and reprojected road edge. In practice, the reprojection accuracy is vital for BEV algorithm training. The flaws of the road surface annotation generation pipeline mainly cause such reprojection inconsistency. Notably, the HD map provided by the scenes dataset is annotated in 2D satellite images. The Global Navigation Satellite System (GNSS) and Real-Time Kinematic (RTK) positioning signals maintain the pose alignment between images and the map. Despite high RTK positioning accuracy, the actual camera pose accuracy is not guaranteed due to synchronization and calibration errors~\cite{zhang2021deep, zhang2022towards}. Thus, the nuScenes HD Map cannot guarantee the quality of annotations in consistency and accuracy. 

\begin{figure}[t!]
    \includegraphics[width=0.5\textwidth]{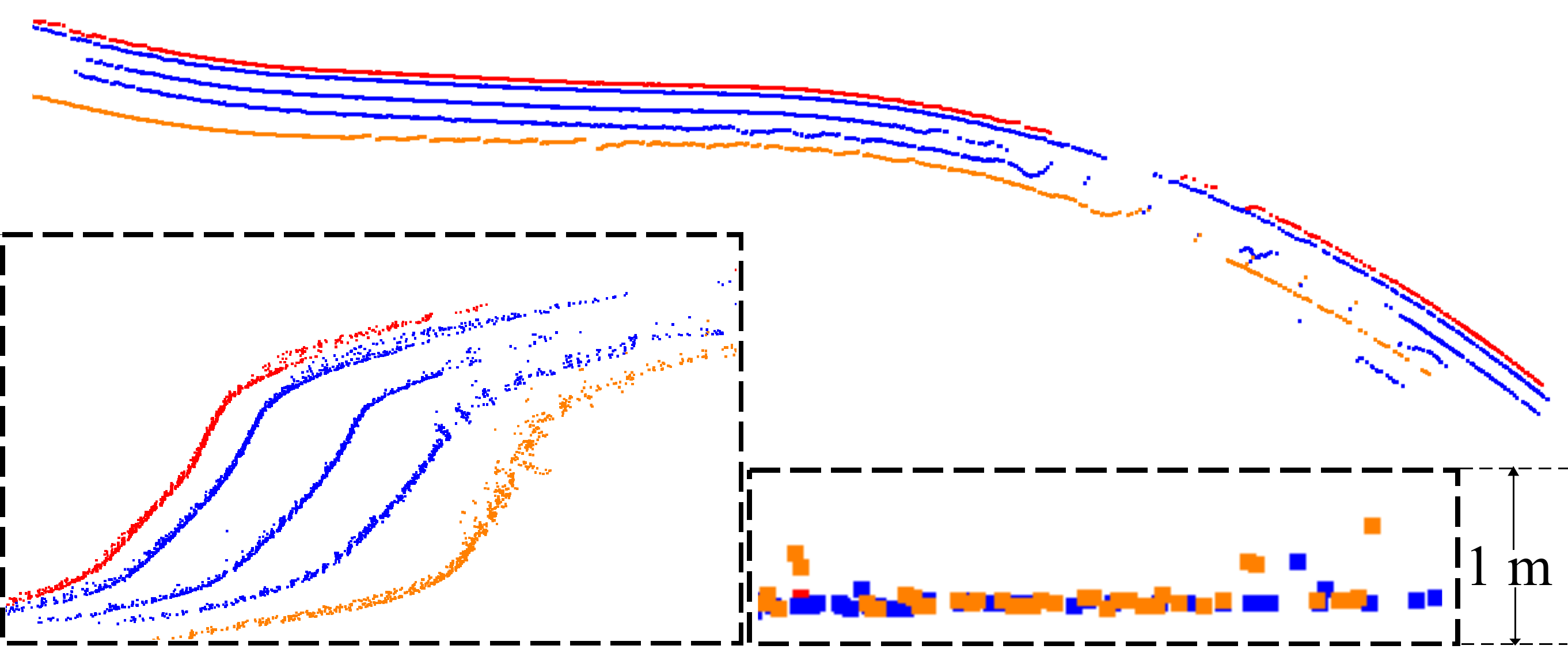}
  \caption{Sparse 3D lane point clouds from the OpenLane V1 dataset. It uses LiDAR point clouds projection to generate 3D lane annotations. The red, blue, and yellow points refer to road teeth, lane dividers, and solid yellow lane marks, respectively. The side view shows that the elevation noises are around the meter level. The zoom-in view also shows that noises are distributed in all directions.}
  \label{fig:openlane_v1}
  \vspace{-1.5em}
\end{figure}

Without HD Map, some works employ LiDAR points for 3D road elements annotation. 
RSRD~\cite{zhao2023rsrd} deploys LiDAR and stereo cameras to reconstruct the road surface with high accuracy focusing on cracks, bumps and potholes, while our work emphasizes more on road structure reconstruction in large scale. OpenLane~\cite{chen2022persformer, li2023topology} and Once-Lane3D \cite{yan2022once} propose to combine 2D image segmentation and LiDAR points to generate 3D lane annotation. 
To do so, the LiDAR points are projected to the image plane to obtain sparse depth. Then, the 2D instance lane segmentation is back-projected to 3D space in camera coordinates from the sparse depth. A filtering algorithm removes the depth noises from inaccurate LiDAR points. Finally, the 3D lane annotations are back-projected into the 3D space using LiDAR points. The advantage is clear: the 2D lane segmentation guarantees geometric accuracy. However, such a method does not impose the spatial-temporal consistency of the 3D lane. The sparse depth could be noisy due to multiple factors, including calibration, synchronization, dynamic objects, and occlusion. Thus, the back-projected 3D lane may not align well with the real one. To verify it, Fig.~\ref{fig:openlane_v1} shows that the 3D lane annotations from LiDAR point projections usually suffer from noises and artefacts. The noises along the elevation direction are around the meter level. 

Unlike the methods above, our proposed CAMA pipeline reconstructs the static map element with surround view images. Without the need for LiDAR and predefined HD Map, the annotation results are more consistent and accurate.

\begin{figure*}[t!]
    \begin{subfigure}{0.3\textwidth}
        \centering
        \includegraphics[width=\textwidth]{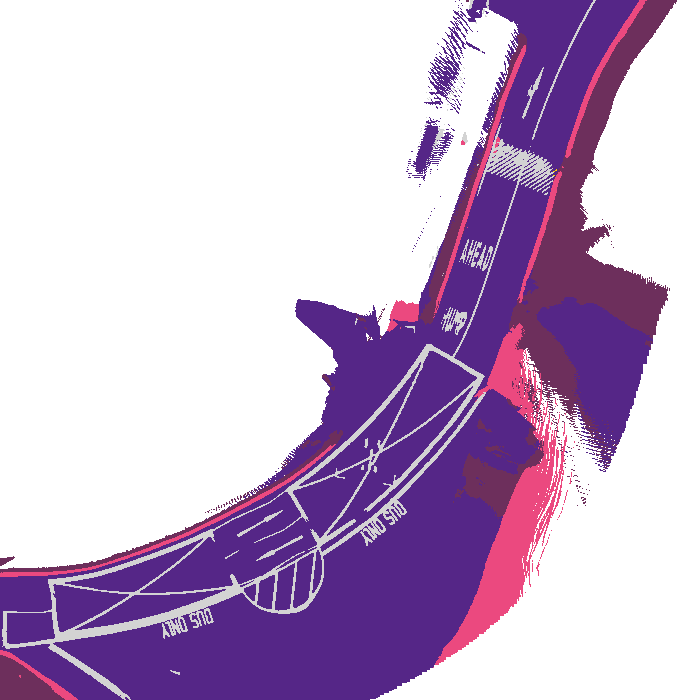}
        \caption{Semantic map.}
    \end{subfigure}
    \hfill
    \begin{subfigure}{0.3\textwidth} 
        \centering
        \includegraphics[width=\textwidth]{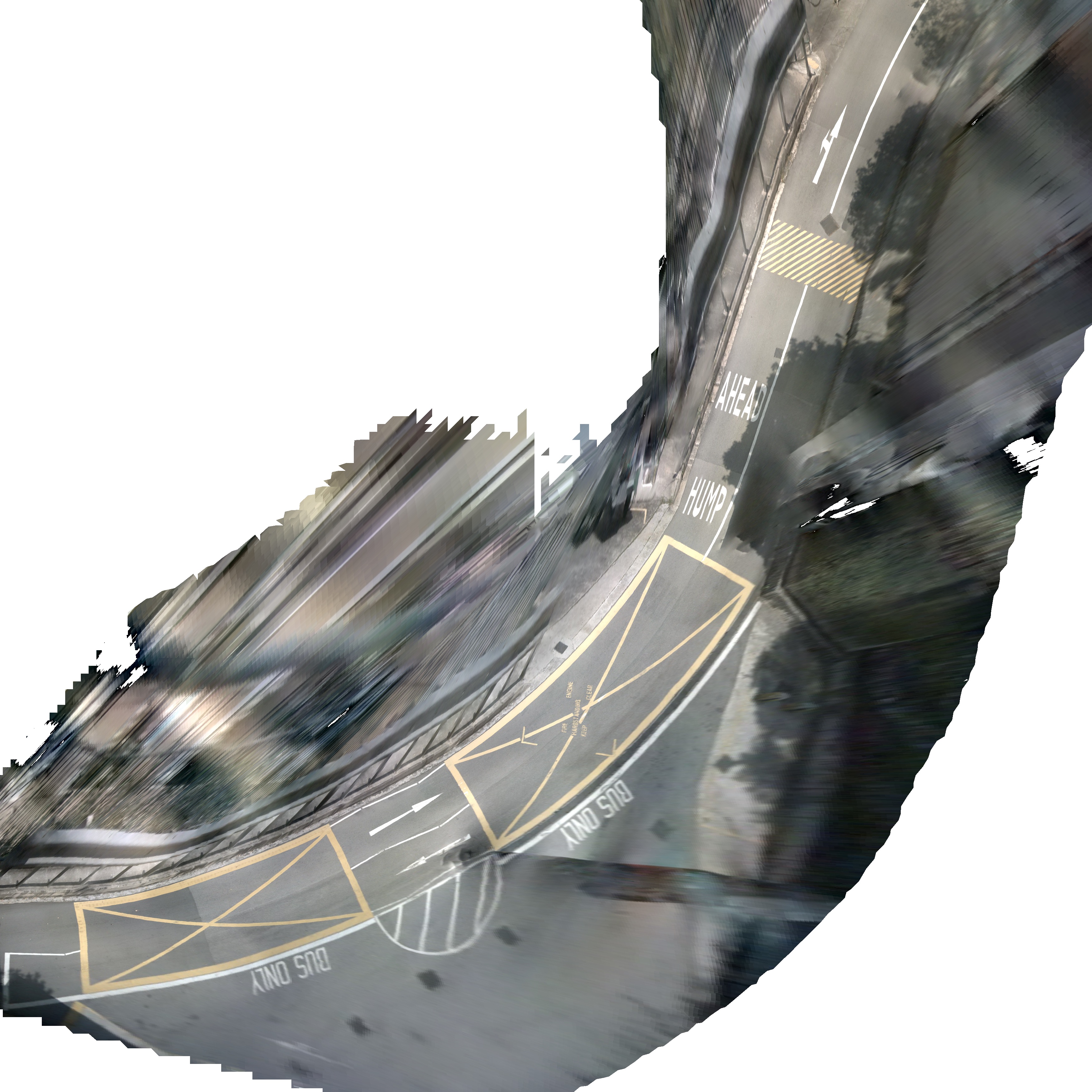}
        \caption{Photometric map.}
    \end{subfigure}
    \hfill
    \begin{subfigure}{0.3\textwidth} 
        \centering
        \includegraphics[width=\textwidth]{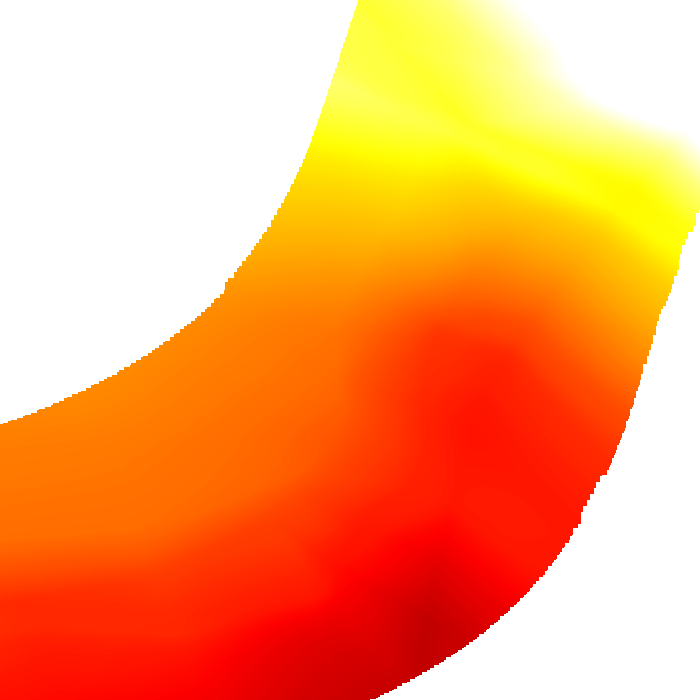}
        \caption{Elevation.}
    \end{subfigure}
    
    \caption{Reconstructed HD Map of scene-0828 from nuScenes using our proposed method. (a) Semantic map in BEV, purple, pink, and white correspond to road surface, road teeth, and lane marking, respectively. (b) Photometric map in BEV. (3) Elevation visualization in hotmap, brighter indicates higher.}
    \label{fig:bev_result}
\vspace{-1.5em}
\end{figure*}

\section{APPROACHES}
\label{sec:approach}

Fig.~\ref{fig:whole_pipeline} details the pipeline of our proposed CAMA. It mainly consists of scene reconstruction and road element vector annotation. The first part is fully automatic, while the second is addressed semi-automatically based on a human-in-the-loop fashion. Mainly, the offline map auto-annotation model~\cite{chen2023vma} is first employed, followed by verification and modification by human annotators. Our pipeline is a vision-centric approach, as the inputs include surround view images and auxiliary sensors (wheel odometry, GNSS, and IMU). Since CAMA guarantees all 3D elements and their correspondence to 2D images, the 3D-2D correspondence and reprojection accuracy are also insured (even improved) without LiDAR.

\subsection{Scene Reconstruction}
As presented in Fig.~\ref{fig:whole_pipeline}, our scene reconstruction comprises three parts (see ): WIGO, SfM, and RoMe.

\subsubsection{WIGO}
We propose to use a Wheel-IMU-GNSS odometry (WIGO) algorithm to combine sensor reading with pose graph optimization. Following VIWO~\cite{lee2020visual}, we combine GNSS signals with pose graph optimization~\cite{gtsam, factor_graphs_for_robot_perception} to ensure global localization and accurate scale. The IMU and wheel sensor provide relative pose constraints between consecutive frames. The WIGO algorithm gives a descent global 6-DoF (Degree of Freedom) pose with a real-world scale. The scale information mainly comes from the GNSS observations. The WIGO results are used as inputs of the Structure from Motion (SfM) pipeline illustrated below. 

\subsubsection{SfM}
Inspired by COLMAP~\cite{schoenberger2016sfm, schoenberger2016mvs}, we introduce an efficient SfM method for whole scene reconstruction. To achieve higher accuracy and efficiency toward self-driving challenges, we optimize COLMAP from five aspects below:
\begin{itemize}
    \item \textbf{Initialization}: The incremental SfM may suffer substantial computation burdens when dealing with large-scale reconstruction (e.g., 300m x 300m areas and thousands of images). In real-world driving scenarios, the pose prior to each image can be easily obtained from GPS and other sensors (e.g., IMU and wheel odometry). Motivated by this, we propose an Odometry Guided Initialization (OGI) for SfM. Specifically, the WIGO pose is transformed into the front camera coordinates with extrinsic. Given the initial poses, the incremental SfM can be replaced with the spatial-guided SfM. The images are initialized with poses from WIGO to conduct triangulation, followed by a series of iterative bundle adjustments (BA).
    \item \textbf{Matching}: The cost of feature matching grows exponentially in traditional exhaustive SfM. Sequential matching may alleviate this issue but bring new problems with matching missing for multiple driving clips. We propose homography-guided spatial pairs (HSP) for the specific driving scenario to balance matching recall and efficiency. Specifically, with the help of WIGO poses, the potential matched image pairs can be filtered by the visual cone overlap between images. Furthermore, as for a self-driving application, all the cameras have an approximate extrinsic to the ground plane. The visual overlap between images can be further filtered by applying a homography transform with respect to the ground plane to emphasize the importance of the road surface area.
    \item \textbf{Feature point}: To further boost the robustness of our pipeline in extreme illuminance and weather conditions, we train a feature point extraction network, SuperPoint~\cite{detone2018superpoint}, on our dataset and pay extra attention to the road surface.
    \item \textbf{BA}: We modify the bundle adjustment part to balance efficiency and accuracy. An iterative BA strategy is applied with a triangulation points filter. The inaccurate points can be removed from the SfM sparse model during the iterations.
    \item \textbf{Rigid prior}: We employ rigid BA to replace the ordinary BA process in COLMAP. The rationale is that multiple cameras on a vehicle are attached to the vehicle's body and can be regarded as mounting on a rigid body. Applying rigid BA directly not only improves the overall efficiency of the pipeline but also increases robustness.
\end{itemize}

With the above optimizations, we achieve roughly five times efficiency boost and 20\% robustness (success rate) improvements for self-driving datasets (see Section~\ref{sec:exp} for details). The accurate 6-DoF poses and corresponding sparse 3D points generated by SfM models are used as input for RoMe to reconstruct road surface mesh (Section~\ref{s:approach:rome}). It should be noted that for the nuScenes dataset, we applied our pipeline by clip (or scene as the term by nuScenes) to get annotation results. The difference is that reconstructing by clips may result in a smaller reconstruction area. Although some scenes in the nuScenes dataset have geographical overlap, the proportion is small, so we neglect these factors.

\subsubsection{RoMe}
\label{s:approach:rome}
We extend our previous work RoMe~\cite{mei2023rome} for road surface mesh reconstruction. The extensions can be roughly divided into three parts:
\begin{itemize}
    \item \textbf{Surface points}: An off-the-shelf 2D segmentation network~\cite{cheng2022masked} is applied to get road and lane segmentation masks. Combined with the sparse SfM models, semantic sparse point clouds can be recovered. The sparse road surface point clouds are then extracted. To further increase the robustness when the SfM points are too sparse, we also sample additional points based on ego-pose and the extrinsic between the cameras to the ground. This approach improves the overall quality of the road surface initialization. 
    \item \textbf{Elevation}: An elevation MLP (Multilayer perceptron)~\cite{gardner1998artificial} is trained with the sparse road surface (3D point clouds) produced in the previous step. 
    \item \textbf{Semantics}: The semantic labels and photometric features are trained by supervising the original images and their corresponding 2D segmentation results.
\end{itemize}

In practice, the elevation MLP could also be optimized and refined in the third part for better consistency among geometry and photometric features. Ultimately, the highly accurate 3D road surface mesh can be obtained. Note that this 3D road surface mesh and elevation maps could also be represented as 2D BEV images and an elevation image as shown in Fig.~\ref{fig:bev_result}. Such representation is fed into the next stage: Map Annotation (Section~\ref{s:approach:annotation}).

\subsection{Map Annotation}
\label{s:approach:annotation}
We extend VMA (Vectorized Map Annotation System)~\cite{chen2023vma} for initial map annotation. VMA is an offline map auto-annotation framework based on MapTR~\cite{liao2022maptr}. The input is the concatenated 2D BEV semantic photometric images, while the output is the vectorized representation of road surface elements (e.g., lane dividers, road teeth, crosswalks, etc.). We propose to use two-layer inputs since they can improve the VMA reasoning ability, especially when classifying the type of lane dividers. Note that this process is still in 2D BEV space. Once we obtain 2D representations, an elevation map is combined to lift the 2D vectors into actual 3D road surfaces and vectors.

In practice, the VMA results are used as priors to accelerate the manual annotation process. While the labeled data is accumulated, the manual works are converted from labeling to verification and minor modification. Still, the entire reconstruction and annotation workflow runs automatically.

\begin{figure}[t!]
    \label{algorithm:sre}
    \includegraphics[width=0.48\textwidth]{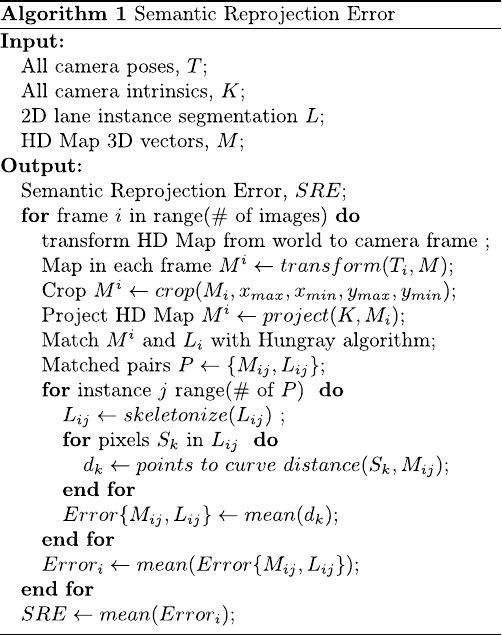}
\end{figure}

\section{EXPERIMENTS}
\label{sec:exp}
The nuScenes dataset is one of the most actively used datasets for online HD Map construction~\cite{caesar2020nuscenes}. It has been widely used to validate recently proposed BEV perception algorithms~\cite{li2022hdmapnet, liu2023vectormapnet, liao2022maptr, liao2023maptrv2}. Other datasets like OpenLaneV2~\cite{chen2022persformer} are also built upon the origin nuScenes dataset. Thus, we validate our proposed framework on the nuScenes dataset to generate optimized annotations.

\begin{table}[t!]
\normalsize
\centering
   \caption{Quantitative comparison between the original nuScenes HD Map and our proposed nuScenes-CAMA HD Map. We evaluate the Semantic Reprojection Error (SRE), Precision, and Recall.}
   \begin{tabular}{c c c c c}
   \hline
            &  SRE $\downarrow$ & Precision $\uparrow$ & Recall $\uparrow$  & $F_1 \uparrow$ \\ \hline
   nuScenes &  8.03           & 0.59               & 0.51             & 0.54               \\ \hline
   CAMA (Ours)   &  \textbf{4.73}  & \textbf{0.87}      & \textbf{0.54}    & \textbf{0.66}      \\ \hline
   \end{tabular}
   \label{table:comparison}
\end{table}

\subsection{Dataset samples}
The nuScenes dataset contains 1000 scenes, each containing 15 seconds of driving data. There are 28130, 6019, and 6008 samples for training, validation, and testing. Our proposed annotation is a subset of the nuScenes dataset, with 14707, 3583, and 3794 samples for training, validation, and testing. Since our method reconstructs each scene independently, the area around each scene's first and last frames can not be fully reconstructed due to a lack of observations from neighboring frames. The head and tail frames are dropped according to the driving distance to guarantee the coverage of our annotations on training and testing sample ranges. To fairly compare our proposed annotation with the original nuScenes dataset, we subsample the nuScenes dataset according to our annotation frames accordingly. We named them nuScenes-sub and nuScenes-CAMA in the experiments.

\begin{table*}[t!]
\normalsize
\centering
   \caption{Comparison of different MapTR model evaluation results on the nuScenes validation dataset. The training and validation use the same kind of annotation for each experiment.
   We also evaluate the reprojection accuracy (SRE) and consistency (precision, recall and $F_1$ score) for the prediction results of different models.}
   \begin{tabular}{c|c c |c c c c |c c c c c c}
   \hline 
    Exp. \#   &  annotation  & w/ elevation    & $AP_{ped} $ & $AP_{divider} $ & $AP_{boundary} $ & mAP $$    & SRE $\downarrow$ & Precision $\uparrow$ & Recall $\uparrow$  & $F_1 \uparrow$  \\ \hline
    \# 1      &  nuScenes    &                 & 50.8        & 45.8            & 55.4             & 50.7      &  8.43            & 0.51                 &  0.37              &  0.42 \\ \hline
    \# 2      &  CAMA        &                 & 47.2        & 41.2            & 36.4             & 41.6      &  6.68            & 0.72                 &  0.52              &  0.60  \\ \hline
    \# 3      &  CAMA        &  \checkmark     & 46.8        & 38.7            & 37.3             & 40.9      &  5.81            & 0.93                 &  0.49              &  0.64  \\ \hline
   \end{tabular}
   \label{table:model_training}
\end{table*}

\subsection{Quantitative Validation}
As introduced in Section~\ref{sec:intro}, the nuScenes dataset provides HD Maps for all scenes, while the annotation does not contain elevation. Reprojecting an HD Map into the image space using the ego-motion and calibration, as presented in Fig.~\ref{fig:reprojection}, we can clearly observe the misalignments between the lane vectors and image space. For quantitative analysis, we propose a metric denoted semantic reprojection error (SRE) to analyze better and compare the reprojection accuracy. The detailed steps are illustrated in Algorithm 1: \textbf{Step 1}: Reproject the 3D annotation vector elements into each 2D image. \textbf{Step 2}: Extract all the instances in image space using an off-the-shelf 2D lane instance segmentation~\cite{jain2023oneformer} and fit polylines for each instance. \textbf{Step 3}: Match the projected elements from Step 1 and the extracted elements from Step 2 using the Hungarian algorithm~\cite{kuhn1955hungarian}. \textbf{Step 4}: Calculate the mean pixel distance for each matched element. 

Table~\ref{table:comparison} compares SRE, precision, recall, and $F_1$ Score between the original nuScenes-sub dataset and our proposed nuScenes-CAMA dataset. We also calculate the precision and recall of the lane divider. Our proposed method achieves 41\% lower SRE, indicating that our reconstructed HD Map and ego-pose have better consistency and higher accuracy. The Precision and Recall are improved 0.28 and 0.03 respectively because our proposed method reconstructs the map based on images, inherently ensuring the correspondence between reconstructed maps and actual image observations.

\subsection{Qualitative Validation}
We observe that the HD Map provided by nuScenes does not always reflect the actual environments captured by cameras. For example, most HD Map annotations show the lane divider between the ego-vehicle and bicycle lanes. However, observe carefully the camera image in Fig.~\ref{fig:reprojection} (a, b): there is no lane marking on the ground in the relevant area. In practice, such inconsistency between the image and HD map annotation increases the training difficulty, particularly the convergence of perception models. An intuitive understanding: there is no lane marking in the image, and the model is likelier to ``memorize" the HD Map according to the nearby environment instead of ``reasoning" the local map based on the observations. Consequently, such inconsistent training data will weaken the potential generalizability of the perception model. Differently, our CAMA can produce accurate (Fig.~\ref{fig:reprojection} d) and consistent (Fig.~\ref{fig:reprojection} f) lane marking, including road teeth, lane divider, and ped-crossing.

\begin{figure}[t!]
    \includegraphics[width=0.50\textwidth]{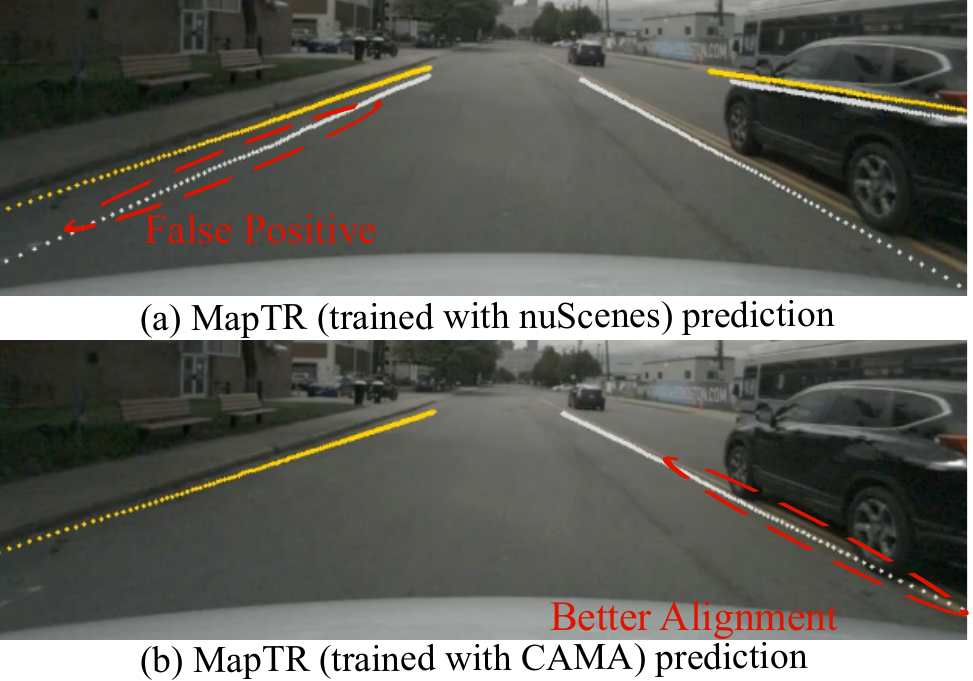}
  \caption{Comparison of the model prediction reprojection. (a) The result reprojection of MapTR model trained with the nuScenes dataset. The red circle shows a false positive prediction, as there is no lane marking on the ground. (b) The result reprojection of MapTR model trained with the CAMA annotation. The red circle shows better reprojection accuracy compared to (a).}
  \label{fig:model_result}
    \vspace{-1em}
\end{figure}

\subsection{Application}
Highly accurate and spatial-temporal consistent HD map annotations are vital for BEV perception algorithm training. To verify the effectiveness of our annotations, we choose MapTRv2 as our baseline model. We removed the auxiliary dense prediction loss in MapTRv2 to reduce the influence of irrelevent supervision signals. We conduct three experiments: 1) MapTRv2 trained on nuScenes-sub dataset; this is set as the baseline. 2) MapTRv2 trained on CAMA but without elevation information. 3) MapTR trained on the CAMA annotation with elevation information. All the experiments are trained 24 epochs with ResNet-50 backbone~\cite{he2016deep}. Table~\ref{table:model_training} details model prediction accuracy and the reprojection metrics, including SRE and $F_1$ score. We can clearly find that models trained with CAMA annotation predict better reprojection accuracy and consistency. The SRE improves from 8.43 to 5.81 by 31 \%, and the $F_1$ score improves from 0.42 to 0.64. We also visually compare the predictions in Fig.~\ref{fig:model_result}. Due to the inconsistent annotation in the nuScenes dataset, the trained MapTR makes false positive predictions (red circle in Fig.~\ref{fig:model_result} a). With our proposed CAMA annotation, the model predictions align better with the image in reprojection (red circle in Fig.~\ref{fig:model_result} b).

\begin{figure}[t!]
    \includegraphics[width=0.48\textwidth]{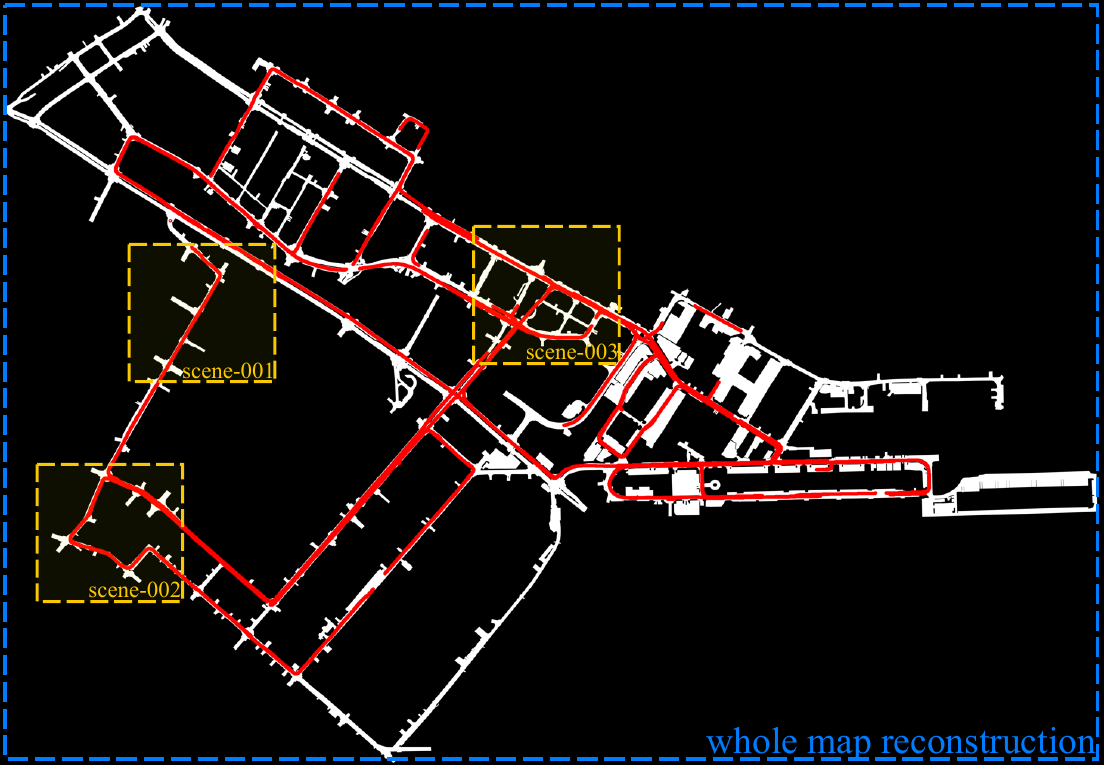}
  \caption{One of the nuScenes map (boston-seaport) visualization. Currently, the CAMA pipeline is conducted by each scene (yellow boxes). In the future, we will release the annotation by each map as presented in the blue dashed box for the whole image. In this way, all the annotations are consistent across different scenes.}
  \label{fig:nuscenes_map}
    \vspace{-1em}
\end{figure}

\section{CONCLUSION}
\label{section:conclusion}

We present CAMA: a vision-centric approach for consistent and accurate static map element annotations. We investigate the critical factors for online HD Map construction and argue that the annotation quality in reprojection accuracy and spatial-temporal consistency is vital for perception algorithm training. Based on this insight, we propose a new baseline for the BEV perception algorithm. Currently, the annotation is a subset of the original nuScenes dataset because we only generate CAMA results for each scene. Fig.~\ref{fig:nuscenes_map} draws the relation between the scenes (yellow boxes) and map (blue box) of the nuScenes dataset. In future works, we will release the CAMA results by map (blue box). 

We choose a decoupled method for reconstructing the physical and logical layers. Initially, it reconstructs the road surface in mesh. Subsequently, it extracts a vectorized representation of lanes in a data-driven method. In contrast, methods like RoadMap~\cite{qin2021light} advocate reconstructing both layers simultaneously. The logical layers are highly complex, with numerous long-tail corner cases, which, we argue, can be more effectively addressed through data-driven approaches. Considering the swift progress of foundation models within computer vision, we anticipate the future development of fully integrated end-to-end reconstruction techniques.



\bibliographystyle{IEEEtran}
\bibliography{ref}

\end{document}